\documentclass[letterpaper, 10 pt, journal, twoside]{IEEEtran}
\IEEEoverridecommandlockouts
\usepackage{graphicx}
\usepackage{overpic}
\usepackage{svg}
\usepackage{xcolor}
\usepackage{bm}
\usepackage[hang,flushmargin]{footmisc}
\usepackage{tabularx}
\usepackage{multirow}
\usepackage{booktabs}
\usepackage{colortbl}
\usepackage{float}
\usepackage{placeins}

\usepackage{array}
\newcolumntype{P}[1]{>{\centering\arraybackslash}p{#1}}  % Define centered p{} column type

\usepackage{amsmath}
\usepackage{amssymb}
\usepackage{amsthm}
\usepackage{siunitx}
\newtheoremstyle{main}
{1em}                                                % space above
{1em}                                                % space below
{\itshape}                                        % bodyfont
{0pt}                                                % indent
{\scshape}                                           % head font
{\\*}                                                % head punctuation
{2pt}                                                % head space
{\thmname{#1}\thmnumber{ #2}: \thmnote{\itshape #3}} % head spec

\usepackage{environ}

\usepackage{algpseudocode}
\usepackage[linesnumbered,ruled,noend]{algorithm2e}
\usepackage[
	activate   = {true},
	protrusion = false,
	expansion  = true,
	kerning    = true,
	spacing    = true,
	tracking   = false,
	auto       = true,
	selected   = true,
	factor     = 2000,
	stretch    = 50,
	shrink     = 20,
]{microtype}
\definecolor{gg}{RGB}{240, 74, 0}

\makeatletter
\let\NAT@parse\undefined
\makeatother
\usepackage[pdfa,colorlinks,bookmarksopen,bookmarksnumbered,allcolors=gg]{hyperref}
\usepackage[noadjust]{cite}
\usepackage[nameinlink,capitalise]{cleveref}
\crefname{line}{line}{lines}
\crefname{figure}{Fig.}{Figs.}
\Crefname{figure}{Fig.}{Figs.}
\crefname{equation}{Eq.}{Eqs.}
\Crefname{equation}{Eq.}{Eqs.}
\crefname{section}{Sec.}{Secs.}
\Crefname{section}{Sec.}{Secs.}
\crefname{definition}{Def.}{Defs.}
\Crefname{definition}{Def.}{Defs.}
\crefname{algorithm}{Alg.}{Algs.}
\Crefname{algorithm}{Alg.}{Algs.}
\crefname{assumption}{Asm.}{Asms.}
\Crefname{assumption}{Asm.}{Asms.}
\crefname{subassumption}{Asm.}{Asms.}
\Crefname{subassumption}{Asm.}{Asms.}
\Crefname{problem}{Problem}{Problems}
\crefname{problem}{Problem}{Problems}

\usepackage{flushend}

\usepackage[inline]{enumitem}
\usepackage{mathtools}

\newcommand{\cgoal}{\mbox{\ensuremath{\mathcal{C}_{\rm goal}}}\xspace}
\newcommand{\cfree}{\mbox{\ensuremath{\mathcal{C}_{\rm free}}}\xspace}
\newcommand{\cstart}{\mbox{\ensuremath{q_{\rm start}}}\xspace}
\newcommand{\workspace}{\mbox{\ensuremath{\mathcal{W}}}\xspace}

\newcommand{\dof}{\textsc{d\scalebox{.8}{o}f}\xspace}
\newcommand{\mopsprm}{\textsc{MOPS-PRM}\xspace}

\newcommand{\ie}{\emph{i.e.},\xspace}

\graphicspath{ {./images/} }

\DeclareMathOperator*{\argmin}{arg\,min}
\newcommand{\calC}{{\cal C}}
\newcommand{\calD}{{\cal D}}

\newcommand{\calN}{{\cal N}}
\newcommand{\calO}{{\cal O}}

\newcommand{\calX}{{\cal X}}

\newcommand{\bfm}{\mathbf{m}}
\newcommand{\bfn}{\mathbf{n}}

\newcommand{\bfq}{\mathbf{q}}

\newcommand{\bfx}{\mathbf{x}}

\newcommand{\bfz}{\mathbf{z}}

\newcommand{\bftheta}{\boldsymbol{\theta}}

\newcommand{\bfpi}{\boldsymbol{\pi}}

\newcommand{\bfphi}{\boldsymbol{\phi}}

\newcommand{\bfI}{\mathbf{I}}

\newcommand{\bfP}{\mathbf{P}}

\newcommand{\bfSigma}{\boldsymbol{\Sigma}}

\newcommand{\bbR}{\mathbb{R}}
\newcommand{\bbS}{\mathbb{S}}

\begin{document}

% Paper headers
\markboth{IEEE Robotics and Automation Letters. Preprint Version. Accepted February, 2026}
{Meng \MakeLowercase{\textit{et al.}}: MOPS-PRM} 

\title{Sampling-Based Motion Planning with Scene Graphs \\
Under Perception Constraints}

% Make room for more info lines in the \author command 
\author{Qingxi Meng$^{1,*}$, Emiliano Flores$^{1,*}$, Thai Duong$^{1}$,  Vaibhav Unhelkar$^{2}$, and Lydia E. Kavraki$^{2}$%
%\thanks{Manuscript received: October, 15, 2025; Revised January, 14, 2026; Accepted February, 18, 2026.}
\thanks{
%This paper was recommended for publication by Editor O. Stasse upon evaluation of the Associate Editor and Reviewers’ comments.
This work was supported in part by NSF 2326390 and NSF CCF-2336612.} 
\thanks{$^{1}$Qingxi Meng, Emiliano Flores, and Thai Duong are with  Department of Computer Science, Rice University, Houston, TX 77005 USA {\tt\footnotesize qm15@rice.edu}}%
\thanks{$^{2}$Vaibhav Unhelkar and Lydia E. Kavraki are with the Department of Computer Science, Rice University, Houston, TX 77005 USA, and also with Ken Kennedy Institute, Rice University, Houston, TX 77005 USA. {\tt\footnotesize vaibhav.unhelkar@rice.edu; kavraki@rice.edu}}%
\thanks{$^{*}$ Equal contribution.} 
\thanks{Digital Object Identifier (DOI): see top of this page.}
}

\maketitle

\begin{abstract}
It will be increasingly common for robots to operate in cluttered human-centered environments such as homes, workplaces, and hospitals, where the robot is often tasked to maintain perception constraints, such as monitoring people or multiple objects, for safety and reliability while executing its task. However, existing perception-aware approaches typically focus on low-degree-of-freedom (\dof) systems or only consider a single object in the context of high-\dof robots. This motivates us to consider the problem of perception-aware motion planning for high-\dof robots that accounts for multi-object monitoring constraints. We employ a scene graph representation of the environment, offering a great potential for incorporating  long-horizon task and motion planning thanks to its rich semantic and spatial information. However, it does not capture perception-constrained information, such as the viewpoints the user prefers. To address these challenges, we propose \textbf{MOPS-PRM}, a roadmap-based motion planner, that integrates the perception cost of observing multiple objects or humans directly into motion planning for high-\dof robots. The perception cost is embedded to each object as part of a scene graph, and used to selectively sample configurations for roadmap construction, implicitly enforcing the perception constraints. Our method is extensively validated in both simulated and real-world experiments, achieving more than $\sim 36\%$ improvement in the average number of detected objects and $\sim 17\%$ better track rate against other perception-constrained baselines, with comparable planning times and path lengths.
\end{abstract}


\section{Introduction}
\label{sec:introduction}

\begin{figure}[!ht]
    \centering
    \includegraphics[width=0.95\columnwidth, trim={0.1cm 0.09cm 0.1cm 0.1cm},clip]{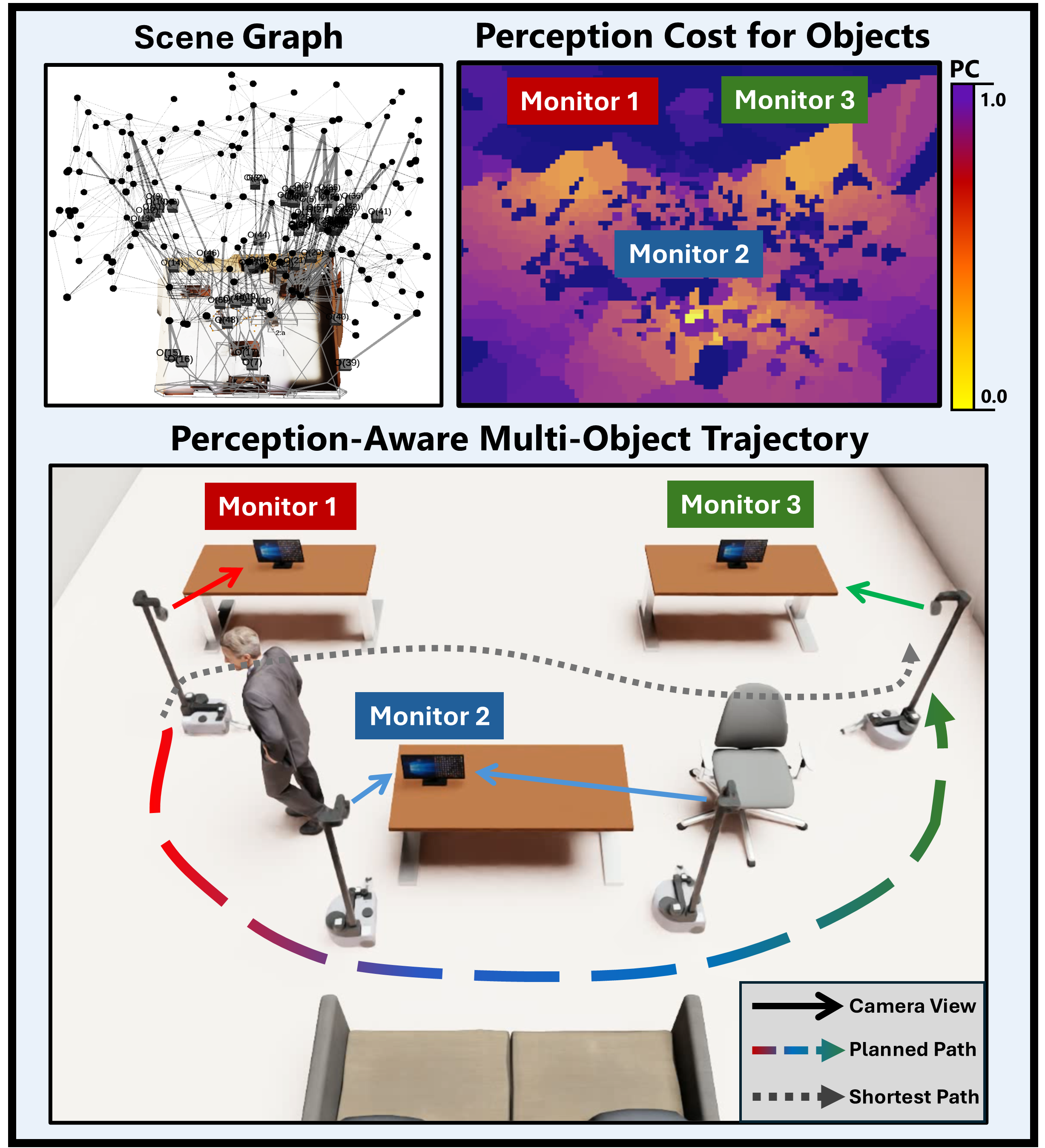}
    \caption{
     An illustration of our perception-aware motion planner that leverages a scene graph embedded with perception costs to generate a trajectory from a start to a goal, while monitoring three objects of interest. The screen of the monitor is the preferable viewpoint in this scenario.}
    \label{fig:illustration_figure}
\end{figure}

\IEEEPARstart{A}{utonomous} robot systems have become more prevalent in a variety of human-centered environments, such as workplaces \cite{bolu2021adaptive}, hospitals \cite{qian2025astrid}, urban areas \cite{liao2023kitti} and homes \cite{shafiullah2023bringing}. A key challenge in such environments is that the robot often has to plan a collision-free trajectory to finish its own tasks, while monitoring other humans or objects of interest along the trajectory. For example, a household robot may need to deliver an item while keeping a person’s face or gestures in view, or a museum patrol robot may navigate around visitors while maintaining visibility of multiple paintings or sculptures. Such perception constraints are essential for monitoring the surrounding objects \cite{falanga2018_pampc, masnavi2024differentiable}, improving the robot's state estimation~\cite{ichter2020_perception,bartolomei2020_perception}, enabling safe navigation~\cite{loquercio2021_learning,song2023_learning}, and mapping and exploration of the environment~\cite{placed2023active_slam_survey}. Therefore, we aim to address the problem of motion planning under perception constraints in this paper.

A common example of perception-aware constraints in motion planning is object tracking and monitoring, where the robot maintains visibility of a single object \cite{falanga2018_pampc, masnavi2024differentiable, meng2025lookleapplanningsimultaneous} or multiple objects \cite{tordesillas2022_panther, zhang2024perception}, that can be either static \cite{falanga2018_pampc, meng2025lookleapplanningsimultaneous} or dynamic \cite{masnavi2024differentiable,tordesillas2022_panther, zhang2024perception} in the environment. Accurate tracking of multiple objects in the environment is particularly useful for collision avoidance, especially with aggressive quadrotor flights \cite{falanga2018_pampc,loquercio2021_learning,song2023_learning}, or in cluttered and dynamic environments~\cite{tordesillas2022_panther,singla2021_memory,masnavi2024differentiable}. Besides objects, several works aim to maintain visibility of visual features for accurate state estimation \cite{ichter2020_perception, bartolomei2020_perception}, such as visual-inertial odometry (VIO) \cite{mourikis2007multi}, which is crucial for robot operations in the wild. Another exciting research direction is active mapping or exploration \cite{placed2023active_slam_survey}, where the robot trajectory is planned to explore the unmapped regions of the environment \cite{bircher2016receding, zhang2020fsmi, asgharivaskasi2023semantic}, either by choosing the next best goal state \cite{bircher2016receding} or by maximizing the information gain of future sensor observations \cite{zhang2020fsmi, asgharivaskasi2023semantic}.

The perception-aware constraints are commonly integrated as a cost, heuristic, or reward function in a motion planning problem, which is in turn, solved by an optimization solver \cite{falanga2018_pampc, tordesillas2022_panther}, a search-based \cite{bartolomei2020_perception} or sampling-based planner \cite{ichter2020_perception, costante2016perception, meng2025lookleapplanningsimultaneous}, or by reinforcement learning \cite{singla2021_memory,song2023_learning}. Although successful in navigation with mobile robots, existing work on perception-aware motion planning largely focuses on settings with simplified robot models or with limited degrees of freedom (\dof). Closely related to our approach, PS-PRM \cite{meng2025lookleapplanningsimultaneous} considers a perception-aware motion planning problem for a high-\dof robot but only monitors a known single object. Extending from monitoring a single object to multiple objects is nontrivial, as the planner must determine how to prioritize and achieve the correct viewpoints of the objects along the trajectory to maximize the overall user-defined perception score. Our work departs significantly from prior work on planning under perception constraints by developing a perception-aware sampling-based motion planner for high-\dof robots, e.g., mobile manipulators, that allows the robot to monitor multiple static objects of interest, stored in a scene graph built from sensor observations, while satisfying the kinematic constraints on the robot configuration.

Recently, metric-semantic maps~\cite{alatise2020review}, such as scene graphs~\cite{chang2021comprehensive}, have emerged as powerful representations that unify geometric, semantic, and topological information for large-scale environments. Scene graphs organize semantic and metric information in a hierarchical structure, capturing relationships across abstraction layers. Scene graphs have been used for task planning and high-level reasoning, in combination with language models, e.g., SayPlan~\cite{rana2023sayplan} and AutoGPT+P~\cite{birr2024autogpt+}, or semantic instructions, e.g., GRID~\cite{ni2024grid} and ConceptGraphs~\cite{gu2024conceptgraphs}, or for explorations, e.g., RoboEXP~\cite{jiang2024roboexp}.

As low-level motion planning requires geometric information and kinematic constraints to ensure the feasibility of a motion plan, recent work has explored the use of scene graphs for both task and motion planning~\cite{ray2024task,dai2024optimal,viswanathan2025spade} in a hierarchical manner.
A ``coarse" task plan is generated at the higher abstraction levels, such as buildings, rooms, or objects, and is then used to guide a local geometric planner at the occupancy level \cite{ray2024task, viswanathan2025spade}, or to generate a heuristic function for a multi-heuristic A* geometric planner \cite{dai2024optimal}. However, these works focus on low-dimensional robot systems, e.g., robot or camera poses, without considering kinematic constraints.

Instead, we develop a sampling-based perception-aware probabilistic roadmap (PRM) planner for high-\dof robots, that integrates the robot's kinematic, geometric, and perception constraints, e.g., multi-object monitoring. The perception constraints are embedded with each object of interest in a scene graph as a perception cost function, which is used to inform the construction of a PRM. Given a robot configuration, the perception cost function describes the perception score of all objects of interest, that can be predefined or approximated by a neural network, pretrained to fit the confidence score of an object detection algorithm such as YOLOE \cite{wang2025yoloe}. For example, a high perception score or low perception cost is given if the camera pose, calculated via forward kinematics, leads to a clear view of multiple objects or humans in the camera image. We develop a perception-aware PRM graph construction by biasedly sampling robot configurations with low perception cost, i.e., high perception score.  Given a start and a goal, an A* search algorithm with our consistent heuristic design returns a robot path on the PRM that balances between the motion cost, representing the path's length or energy, and the perception cost, representing how well the robot can monitor the objects of interest along the path. We extensively validate our approach in both simulation and real-robot experiments. %, 
In summary, we propose a \textbf{M}ulti-\textbf{O}bject \textbf{P}erception-aware \textbf{S}cene-graph-based \textbf{P}robabilistic \textbf{R}oad\textbf{M}ap (\mopsprm) that:
\begin{itemize}
\item augments each object of interest in a scene graph with a learned perception costmap, specifying the preferable configuration regions for multi-object monitoring.
\item constructs a perception-informed PRM on the configuration space of a high-\dof robot by selectively sampling nodes with low perception cost.
\item generates a perception-aware trajectory with an A* search on the perception-aware PRM.
\end{itemize}
\begin{figure*}
    \centering
    \includegraphics[width=\textwidth]{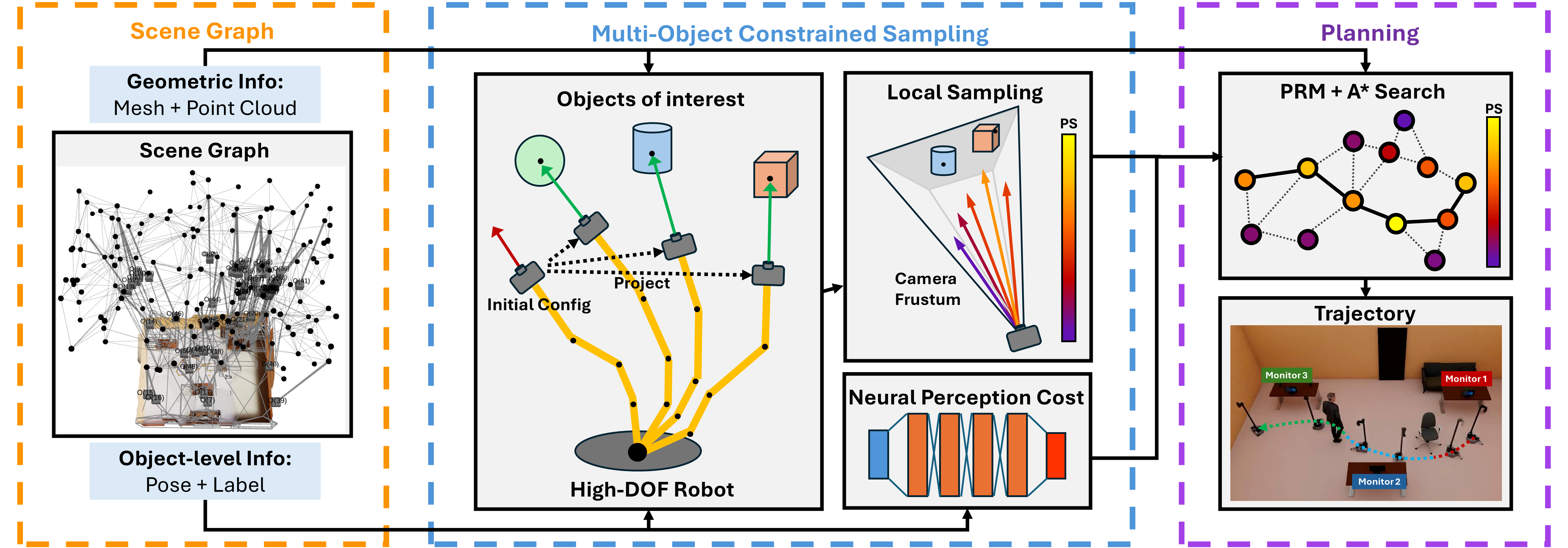}
    \caption{This figure presents the pipeline of our planner. The planner takes the scene graph as input, combining geometric and object-level information with a neural perception cost function to perform multi-object constrained sampling. Sampling is performed (see \cref{sec:method_multiobject}) to construct a PRM, which is searched using A* to generate a trajectory that effectively accomplishes perception tasks involving multiple objects along the path.}
    \label{fig:approach_diagram}
\end{figure*}

\section{Problem Statement}
\label{sec:problem_statement}

We consider a robot with configuration 
$\mathbf{q} \in \mathcal{C} = \mathcal{C}_{\rm free} \cup \mathcal{C}_{\rm occupied} \subseteq \mathbb{R}^n$, 
where $n$ denotes the total number of degrees of freedom, including both the robot’s base and its joints, and $\cfree$ and $\calC_{\rm occupied}$ are the free and occupied spaces, respectively. The robot operates in a workspace $\workspace \subseteq \mathbb{R}^3$ and is equipped with an onboard steerable RGB-D camera, controlled via the robot joints to observe the environment.

The goal of motion planning is to find a collision-free path $\bfpi: [0, 1] \rightarrow \cfree$, from the start configuration $\bfpi(0) = \cstart$ to a goal region $\bfpi(1) \in \cgoal$. Along a path $\bfpi$, the robot also aims to monitor a set $\calO$ of $N$ objects of interest.

A motion cost $c_m(\bfpi)$, defined over the space of all possible paths $\Pi$, assigns a non-negative real number to each path $c_m: \Pi \rightarrow \mathbb{R}_{ \geq 0}$, e.g., its length, energy or control effort. To model the object monitoring constraints, we introduce a perception cost function $f: \cfree \times \calO \rightarrow \mathbb{R}_{\geq 0}
\label{eq:perception_function}$ that assigns each pair of configuration $\bfq \in \cfree$ and object $o \in \calO$ a scalar value, measuring the quality of the observation of $o$ by the onboard camera when the robot is at configuration $\bfq$. A lower perception quality implies a higher perception cost. We next define the overall perception cost at a configuration $\bfq$ as the weighted sum of the object-wise cost:
\begin{equation}
p(\bfq) = \sum_{o \in \calO} w_o \, f(\bfq, o),
\label{eq:perception_config}
\end{equation}
where the weight $w_o$ is a user-defined importance of monitoring object $o$. The cumulative perception cost of a path is:
%the integral of this function along the trajectory,
\begin{equation}
c_p(\bfpi) = \int_0^1 p(\bfpi(t)) \, dt,
\label{eq:perception_path}
\end{equation}
which aggregates the perception cost along the path $\bfpi$. 
Our goal is to find the optimal path $\bfpi^*$ that minimizes the weighted combination of motion and perception cost, \emph{i.e.},
\begin{equation}
\begin{aligned}
    &\bfpi^* = {\argmin}_{\bfpi \in \Pi} \; c_m(\bfpi) + \alpha c_p(\bfpi), \\
    \text{s.t.} \quad &\bfpi(t) \in \cfree \quad\forall t\in [0,1],\\
    &\bfpi(0) = q_{\rm start}, \bfpi(1) \in \cgoal,
\end{aligned}
\label{eq:optimal_path}
\end{equation}
where the weighting factor $\alpha \geq 0$ controls the trade-off between motion and perception cost.

\section{Perception-aware Planner with Multi-Object Monitoring using Scene Graphs}
An overview of \mopsprm is provided in \cref{sec:method_planner} with details on how we develop our perception-informed PRM construction \cref{sec:method_multiobject}, and how we augment a scene graph with the perception cost in 
\cref{sec:method_scenegraph}.

\subsection{MOPS-PRM Planning}
\label{sec:method_planner}
While finding the optimal trajectory in \cref{eq:optimal_path} is challenging due to the high-dimensional configuration space 
of high-\dof robots, our approach instead constructs a probabilistic roadmap (PRM) in the free space 
\cfree and searches for an optimal path $\bfpi$ on the PRM from the start to the goal. 

We define the PRM $G=(V,E)$, where $V$ is the set of $P$ collision-free configurations and $E$ the set of edges connecting them. The edges in $E$ are checked for collisions with obstacles by a validity checker, e.g.,~\cite{thomason2024vamp}, during our PRM construction.
The nodes are generated by a ``perception-aware" sampling scheme (\cref{sec:method_multiobject}), connected to their $k$-nearest neighbors and checked for collision, creating a set of PRM edges (see lines 14-19 of \cref{alg:prm_pseudocode}). Each edge 
$(\bfq_u, \bfq_v)$ is represented by a local motion $\bfpi_{uv}(t), \; \bfpi_{uv}(0) = \bfq_u, \; \bfpi_{uv}(1) = \bfq_v$, which 
depends on the kinematic constraints of the robot, e.g., a Reeds-Shepp curve for the base of a 
non-holonomic mobile manipulator. Each edge $(\bfq_u, \bfq_v)$ is assigned a cost:
\begin{equation}
c(\bfq_u, \bfq_v) = c_m(\bfpi_{uv})+ \alpha \cdot c_p(\bfpi_{uv}),
\label{eq:edge_cost}
\end{equation}
where $c_m(\bfpi_{uv})$ is the edge's motion cost, e.g., the length or total control efforts of $\bfpi_{uv}$, and $c_p(\bfpi_{uv})$ is the edge's 
perception cost from Eq.~\eqref{eq:perception_path}:
\begin{equation}
     c_p(\bfpi_{uv}) = \int_0^1 \sum_{o \in \calO} w_o \, f(\bfpi(t), o)\, dt.
\end{equation}
However, the perception cost $f(\bfq, o)$ for each pair $(\bfq, o)$ is typically unknown in advance for an arbitrary configuration~$\bfq$. Therefore, we approximate the perception cost by a 
neural costmap $f_{\bftheta}(\bfq, o)$ with parameter $\bftheta$, trained on supervised data from an object detector and augmented to each object $o\in\calO$ in a scene graph (see \cref{sec:method_scenegraph}). 
As a result, the edge's perception cost is approximated as:
\begin{equation}
    \label{eq:edge_perception_approx}
     c_p(\bfpi_{uv}) \approx \sum_{k = 0}^{K-1} \sum_{o \in \calO} w_o \, f_{\bftheta}(\bfpi(t_k), o)\,  \delta t, 
     \quad K \geq 2,
\end{equation}
where $t_0, t_1, \ldots, t_K$ are discrete times sampled along the edge with time steps $\delta t = \tfrac{1}{K}$.

After the PRM is constructed, \mopsprm connects the start and goal to the roadmap and applies an A* search~\cite{hart1968formal} with a consistent heuristic to find a path from $\cstart$ to $\cgoal$ for the robot to follow.
Following~\cite{meng2025lookleapplanningsimultaneous}, we define a consistent ``hop"-based heuristic function that lower-bounds the remaining path cost to the goal:
\begin{equation}
h(\mathbf{q}) \;=\; H_{\min} \cdot c^{\min},
\label{eq:heuristic_def}
\end{equation}
where $H_{\min}(\mathbf{q})$ denotes the hop distance, i.e., the minimum number of edges required to reach the goal from the configuration $\mathbf{q}$ obtained via a shortest-path search, and $c^{\min}$ represents the minimum edge cost over the entire roadmap:
\begin{equation}\label{eq:min_edge_def}
c^{\min} = \min_{(u,v)\in E} c(\bfq_u, \bfq_v).
\end{equation}

For any $(u,v)\in E$, a feasible path from $u$ to the goal is to take $(u,v)$ and then follow a shortest path from $v$ to the goal. Thus, we have: $H_{\min}(\bfq_u) \le 1 + H_{\min}(\bfq_v)$.

Therefore, our heuristic function $h(\cdot)$ is consistent:
\begin{align*}
h(\bfq_u) 
     &\le (1+H_{\min}(\bfq_v))\cdot c^{\min}\\
     &\le c(\bfq_u,\bfq_v) + h(\bfq_v).
        &&\text{by \cref{eq:min_edge_def,eq:heuristic_def}}
\end{align*}

This formulation only requires nonnegativity of edge costs, without assuming a specific form such as Euclidean distance for motion cost. It applies broadly, e.g., when $c_m$ is defined as the trajectory length, energy, or control effort, and $c_p$ is a non-negative cost derived from neural perception scores. The weight $\alpha$ controls the tradeoff between the motion and perception costs, e.g., the higher the weight $\alpha$ is, the longer path the A* search might return and vice versa.

The solution returned solves \cref{eq:optimal_path} only for trajectories that lie on the PRM. However, as the number of nodes increases, the solution asymptotically converges to the true optimal trajectory. \mopsprm is illustrated in \cref{alg:prm_pseudocode} with implementation details provided in \cref{sec:implementation}.

\begin{algorithm}[!ht]
\caption{\mopsprm Construction}
\label{alg:prm_pseudocode}
\DontPrintSemicolon
\KwIn{Set of objects of interest $\calO$, neural costmap $f_{\bftheta}$, scene graph $S$, number of nodes $P$}  
\KwOut{Our perception-aware PRM $G=(V,E)$}
$V,E \leftarrow \emptyset$\;

\While{$|V| < P$}{
     \tcp{\small Stage 1: Sampling \& projection}
    $\bfq_0 \leftarrow$ \textsc{Sample}($\cfree$)\;

    $\calC_{\bfq_0} \leftarrow \emptyset$\;
    $\calX_\calO \leftarrow \textsc{ExtractCentroids}(\calO)$\;
    \ForEach{$c$ in $\calX_\calO$}{
      $\bfq_0^c \leftarrow \textsc{Project}(\bfq_0, c)$ (\cref{eq:projection_argmin})\;
      \tcp{\small Stage 2: Local sampling}
      $\calC_{\rm local}^c \leftarrow \textsc{SampleLocal}(\bfq_0^c)$
        
        \ForEach{$\bfq_{\rm loc}^c \in \calC_{\rm local}^c$}{
        \tcp{\small Check field of view (FOV)}
            \If{$c \in \textsc{FOV}(\bfq_{\rm loc}^c)$}{
              $\calC_{\bfq_0} \leftarrow \calC_{\bfq_0} \cup \{\bfq_{\rm loc}^c\}$\;
            }
        }
        }

    \tcp{\small Neural costmap \& selection}
    $\bfq_{\rm node} \leftarrow \arg\min_{\bfq \in \calC_{\bfq_0}}\, p(\bfq)$ (\cref{eq:approx_perception_cost})\;
    $V \leftarrow V \cup \{\bfq_{\rm node}\}$\;
    \tcp{\small Edge connection}
    \ForEach{$\bfq_{\rm near} \in \textsc{NearestNeighbors}(\bfq_{\rm node})$}{
      $\bfpi_{loc} \leftarrow \textsc{LocalMotion}(\bfq_{\rm node},\, \bfq_{\rm near})$\;
      \If{$\textsc{IsCollisionFree}(\bfpi_{loc},\,S)$}{
        $c_p \leftarrow \textsc{PerceptionCost}(f_{\bftheta}, \calO)$ (\cref{eq:edge_perception_approx})\;
        $c \leftarrow c_m(\bfpi_{loc}) + \alpha\,c_p(\bfpi_{loc})$ (\cref{eq:edge_cost})\;}
        $E \leftarrow E \cup \{(\bfq_{\rm node}, \bfq_{\rm near}, c)\}$\;
    }
}
\Return{$(V,E)$}\;
\end{algorithm}

\subsection{Perception-aware Sampling}
\label{sec:method_multiobject}
An important subroutine in \mopsprm is to sample a set of nodes for roadmap construction. 
As the configuration space is high-dimensional, it is beneficial to bias the sampling process 
towards regions with low perception cost. Given a sampled configuration $\bfq_0 \in \cfree$, we would like to find a nearby $\bfq^*$ that minimizes the perception cost:
\begin{equation}
\label{eq:ideal_proj}
\bfq^* = {\argmin}_{\bfq \in \calC_{free}} \; 
\Big( p(\bfq) + \lambda \|\bfq - \bfq_0\|_2^2 \Big),
\end{equation}
where $p(\bfq)$ is the perception cost in \cref{eq:perception_config} and the distance 
$\|\bfq - \bfq_0\|_2^2$ is a regularization term with coefficient~$\lambda$ to penalize 
large deviation from $\bfq_0$. However, solving \cref{eq:ideal_proj} exactly is challenging for our PRM construction, as the perception cost is approximated by:
\begin{equation} \label{eq:approx_perception_cost}
    p(\bfq)~\approx~\sum_{o \in \calO} w_o \, f_{\bftheta}(\bfq, o),
\end{equation}
with a nonlinear neural costmap $f_{\bftheta}(\bfq, o)$. Instead, we introduce a perception-aware local sampling scheme that empirically approximates $\bfq^*$ in two stages, as outlined in \cref{alg:prm_pseudocode}. The first stage (lines 1--7 of \cref{alg:prm_pseudocode}) projects the sampled configuration $\bfq_0$ on a constrained manifold, where the camera pose points toward the objects. The second stage (lines 8--12) performs local sampling around each projection and selects the configuration with the lowest perception cost.

In the first stage, given the sample $\bfq_0$, we calculate the camera pose via forward 
kinematics, and generate viewpoint candidates by projecting the camera optical axis toward 
a set $\calX_\calO$ of $\big(N~+~\tbinom{N}{2}~+~1\big)$ desired centroids, consisting of 
the centroid of each object $o \in \calO$, the 
centroid of each object pair, and the centroid of all objects collectively. This captures common cases where potentially the best viewpoint 
either focuses on observing a single object, all objects, or the transition between a pair of objects. 

After experimentation, we observed that lower perception costs are obtained when the camera’s optical axis is aligned with the object centroid (as illustrated in the “Multi-Object Constrained Sampling” block in \cref{fig:approach_diagram}). Let $\mathbf{m}(\mathbf{q}) \in \mathbb{R}^3$ denote the camera center, 
$\mathbf{z}(\mathbf{q}) \in \mathbb{S}^2$ the unit optical axis, 
and $\mathbf{x}_c \in \mathbb{R}^3$ the 3-D coordinates of the desired centroid 
$c \in \mathcal{X}_{\mathcal{O}}$. We define the lateral (image-plane) projection residual as
\begin{equation}
\bfphi(\bfq,c) := 
\big(\bfI_3 - \bfz(\bfq)\bfz(\bfq)^\top\big)\,
\big(\bfx_c - \bfm(\bfq)\big) \in \bbR^3,
\end{equation}
where $\bfI_3$ is the $3\times 3$ identity matrix~\cite{ma2004invitation}. 
The matrix $\bfP(\bfq) = \bfI_3 - \bfz(\bfq)\bfz(\bfq)^\top$ 
is the orthogonal projector onto the tangent plane of $\bbS^2$ at $\bfz(\bfq)$, i.e., 
$\bfz(\bfq)^\top \bfphi(\bfq,c)=0$ and the residual  
$\bfphi(\bfq,c)$ has only two degrees of freedom corresponding to the lateral error in the image plane. 
We then project $\bfq_0$ onto a configuration, whose camera pose aligns with the centroid $c \in \calX_\calO$ by solving:
\begin{equation}
\label{eq:projection_argmin}
\begin{aligned}
\bfq_0^{c} &= \argmin_{\bfq\in\bbR^k}\;
 \|\bfphi(\bfq,c)\|^2 + \lambda \|\bfq - \bfq_0\|^2_2 \\[2pt]
\text{s.t.}\quad
& \bfq \in \calC_{\mathrm{free}},\; \|\bfq - \bfq_0\|_2 \le \rho,
\end{aligned}
\end{equation}
%\bfq_{\min} \le \bfq \le \bfq_{\max}, 
where $\lambda \ge 0$ is a regularization weight that encourages the solution to be close to $\bfq_0$, 
and $\rho > 0$ is an optional trust-region radius restricting the projection to
a ball around $\bfq_0$. The parameters $\lambda$ and $\rho$ allow us to balance between the PRM coverage of the configuration space and biased sampling towards regions with low perception cost. For a large $\lambda$/small $\rho$, the projected point $\bfq_0^c$ stays close to the uniformly sampled $\bfq_0$, encouraging more even coverage of the configuration space. For a small $\lambda$/large $\rho$, the projected point $\bfq_0^c$ tends to be biased towards regions with low perception cost. The projection problem \cref{eq:projection_argmin} can be solved efficiently via gradient descent, e.g., using an L-BFGS-B solver~\cite{zhu1997lbfgsb}. If the optimization does not converge within the iteration limit, we discard the sample $\bfq_0$ and obtain a new one.

In the second stage, we sample $M$ configurations 
$\{\bfq_{0(i)}^{c}\}_{i = 1}^M$ around each projected $\bfq_0^c$ by adding a zero-mean Gaussian noise $\bfn \sim \calN(\bf0, \bfSigma)$
so that the corresponding camera's field of view (FOV) will still capture the centroid $c \in \calX_\calO$. This process generates a set of $\big(N~+~\tbinom{N}{2}~+~1\big)(M+1)$ candidates: $ \calC_{\bfq_0}~=~
\left\{ \bigcup_{c\in \calX_\calO} \{\bfq_{0(i)}^{c}\}_{i = 1}^M \cup \{\bfq_0^c\}\right\}$. The perception cost of all candidates in $\calC_{\bfq_0}$ is calculated via 
\cref{eq:approx_perception_cost} efficiently in parallel using the neural costmap 
$f_{\bftheta}(\bfq, o)$ (see~\cref{sec:method_scenegraph}). 
The candidate with the lowest perception cost: $\bfq_{\mathrm{node}} = \argmin_{\bfq \in \calC_{\bfq_0}} p(\bfq)$, is added to our perception-aware PRM (lines 12-13 in \cref{alg:prm_pseudocode} and ``Local Sampling'' block in \cref{fig:approach_diagram}).

\subsection{Embedding Perception Costs in Scene Graphs}
\label{sec:method_scenegraph}
Many perception constraints can be characterized by a scalar score $s$, such as the confidence value output $s \in [0,1]$ of an object detection model, \ie YOLOE~\cite{wang2025yoloe}, which can be converted to a %user-defined
perception cost $l$, e.g., $\ell = 1 - s$ or $\ell = 1/s$.
To efficiently query the cost during PRM construction, we train a neural network $f_{\bftheta}(q,o)$ that predicts the perception cost of a pair of configuration $\bfq \in \cfree$ and object $o \in \calO$. The network $f_{\bftheta}(q,o)$ is pre-trained on a wide range of common objects, e.g., representative objects in a home or a hospital, and can be used across different environments. We consider the robot's forward kinematics as a non-trainable first layer of the neural network, calculating the camera pose from the configuration $\bfq$. The input of the second layer is the relative pose between the robot’s onboard camera and the object $o$, together with an encoding of the object’s semantic class, such as \emph{monitor} or \emph{human}. This is followed by a neural network, such as a multi-layer perceptron (MLP), that outputs an estimate of the perception cost $f(\bfq, o)$. 

The training dataset $\calD~=~\{(\bfq_i, o_i, s_i)\}_{i = 1}^D$ is generated using a task-specific perception model, which provides the perception score $s_i$ for each robot-object pair $(q_i, o_i)$. The neural costmap $f_{\bftheta}$ is trained via supervised learning to fit the dataset $\calD$, enabling batched parallel evaluation of perception costs. In practice, not all objects in the scene graph are assigned a perception costmap; only those important to monitor are given this attribute. 
\section{Experimental Results}
\label{sec:experiment}

In our experiments, we verify the effectiveness of our multi-object perception-aware scene-graph-based PRM with simulated and real-robot experiments using the Hello Robot's Stretch $2$~\cite{kemp2022design} and Isaac Sim~\cite{nvidia2022isaacsim} for simulation and visualization.  The Stretch 2 is a high-\dof mobile manipulator: in our experiments, we control its differential-drive, non-holonomic base (3 DoF) together with the pan–tilt joints of the onboard camera, introducing nontrivial kinematic constraints, leading to a challenging perception-aware planning problem. All experiments were conducted on an Intel i7-12700K CPU and an NVIDIA GeForce RTX4090 GPU.

\subsection{Implementation Details} \label{sec:implementation}

While \mopsprm planner can admit different motion and perception costs, and different forms of scene graphs, we present the specific implementation choices that we used.

The motion cost $c_m(\boldsymbol{\pi})$ of a path $\boldsymbol{\pi}$ 
is computed as the sum of the Euclidean distances of all individual edges along the path. 
For the A* heuristic in \cref{eq:heuristic_def}, we use the Euclidean distance between 
the current configuration $\mathbf{q}$ and the goal as the motion component. Meanwhile, the perception cost label $\ell$, used to train our neural costmap in \cref{sec:method_scenegraph}, is chosen as a quadratic function $(1-s)^2$, where $s$ is the confidence score provided by the object detector YOLOE~\cite{wang2025yoloe}. We chose this perception cost to emphasize on higher confidence scores, guiding the planner to favor views that yield more reliable detections. As the motion cost and the perception cost have different units and ranges, we normalize both costs to the range $[0,1]$ using their minimum and maximum values over the entire roadmap, easing parameter tuning for the weight~$\alpha$.

For perception-aware sampling in~\cref{sec:method_multiobject}, an \mbox{L-BFGS-B } solver~\cite{zhu1997lbfgsb} is used to solve \cref{{eq:projection_argmin}} with parameters $\rho =0.05$, $\lambda = 0.3$, and the maximum iterations set to~100. For the local sampling function in the second stage, we choose $M=5$ and use a Gaussian noise $\bfn \sim \calN(\bf0, \bfI)$. For the PRM nearest-neighbor selection in line 14 of \cref{alg:prm_pseudocode}, the number of neighbors is set to be 5, empirically balancing graph connectivity and computational efficiency.

\begin{figure}[t]
    \centering
    \includegraphics[width=\columnwidth]{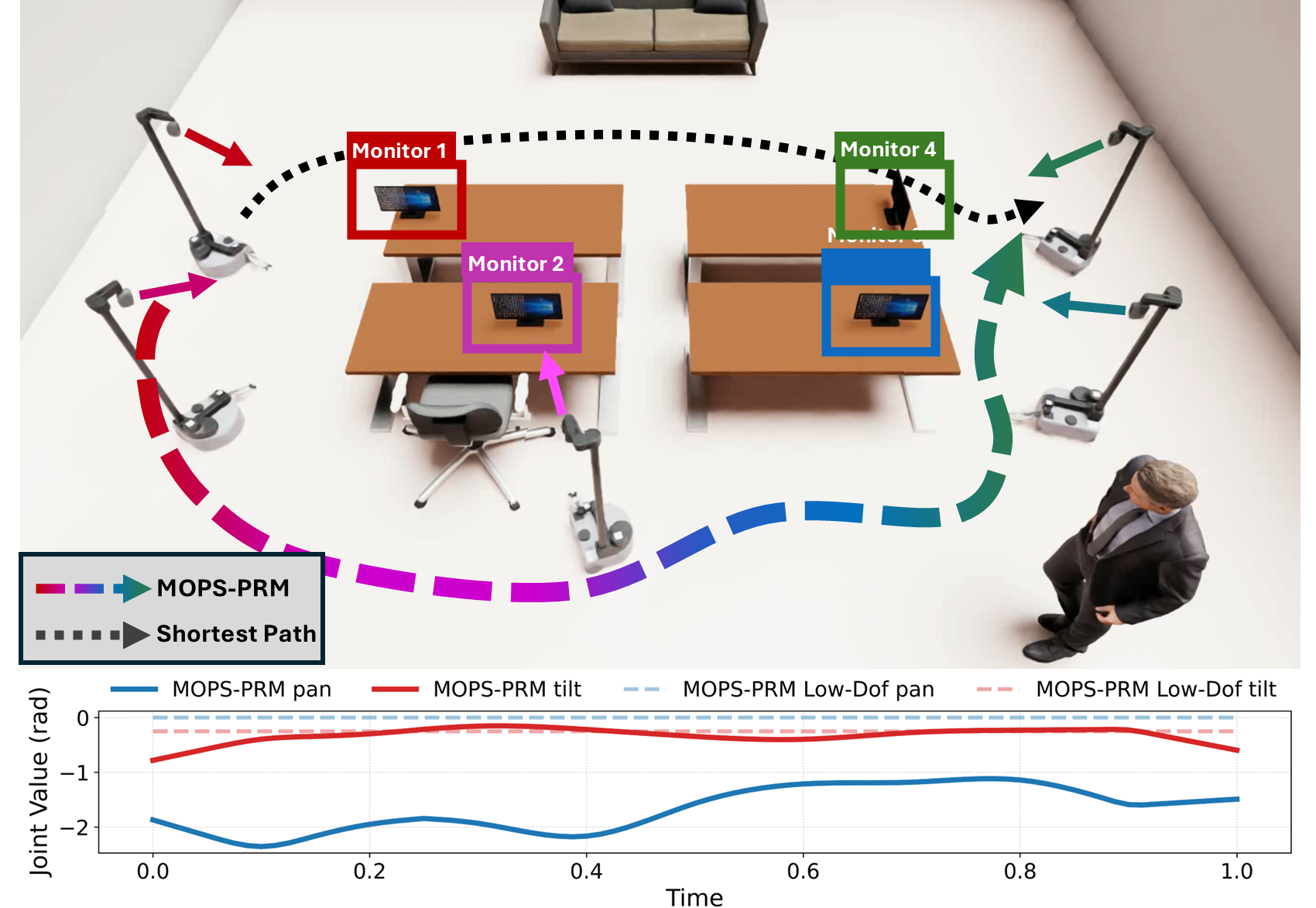}
    \caption{
    In our simulated benchmarks, the robot moves from a start to a goal in an office environment while monitoring the four screens of the monitors placed on the table. The robot takes the longer path to observe the monitors, where the arrows illustrate the camera orientations. The bottom plot shows the camera pan-tilt joint angles along the trajectory.
    }
    \label{fig:simulation_experiment}
\end{figure}

\begin{figure*}[t]
    \centering
    \includegraphics[width=\textwidth]{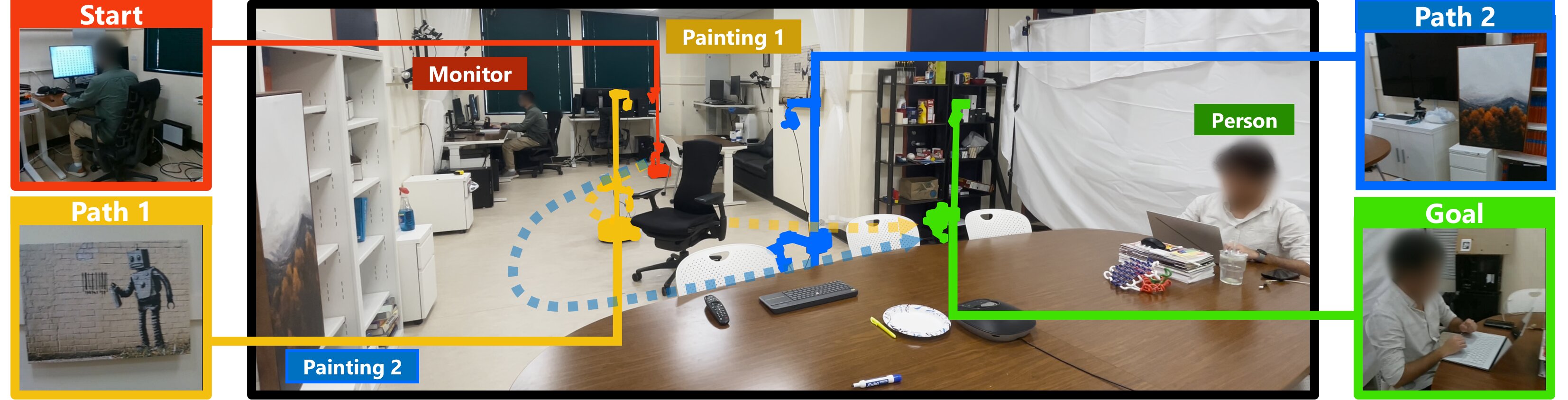}
    \caption{In this real-robot experiment, the robot plans two different paths with the same start (shown in red) and goal (shown in green) based on a user-specified importance of the two paintings: the yellow path prioritizes painting 1, while the blue path prioritizes painting 2. Both paths start by observing a human with a monitor and end by looking at another human sitting at the table while the middle sections of the paths differ as they prioritize observing different paintings.}
    \label{fig:real_robot_dynamic}
\end{figure*}

\begin{table*}[t]
\caption{This table compares the performance of our method with three baselines introduced in \cref{sec:experiment}. For planning metrics, \textit{Build Time},  \textit{Plan. Time}, and \textit{Path Len.} denote the average PRM construction,  planning time, and the average path length, respectively. 
For perception metrics, \textit{Avg. Det. Obj.}, \textit{Track Rate}, \textit{Avg. Conf.}, and \textit{Scaled Avg. Conf.} denote the average number of detected objects per frame, the tracker update success rate, the average detection confidence score, and the confidence score scaled by the average number of detected objects, respectively.}
\centering
\resizebox{\textwidth}{!}{
\begin{tabular}{lcccc|cccc}
\toprule
 & \multicolumn{4}{c}{Motion Planning} & \multicolumn{4}{c}{Perception} \\
\cmidrule(lr){2-5} \cmidrule(lr){6-9} 
 \multirow{2}{*}{Method} & Cost Fun. & Build Time & Plan. Time & Path Len. & Avg. Det. Obj. & Track Rate & Avg. Conf. & Scaled Avg. Conf. \\
 & & (s) & (s) & & (0--4) & (0--1) & (0--1) & (0--4) \\
\midrule
\rowcolor{gray!1}Closest-Object Low-\dof & Distance& 5.0 & 0.22 & 16.63 & 0.44 & 0.72 & 0.45 & 0.20 \\
\rowcolor{gray!5} Closest-Object & Distance& 5.0 & 0.08 & 15.66 & 0.50 & 0.65 & 0.42 & 0.21 \\
\rowcolor{gray!10} Lowest-Cost-Object & Neural & 30.0 & 0.95 & 19.07 & 1.12 & 0.76 & 0.48 & 0.54 \\
\rowcolor{gray!15} \mopsprm (Ours)  & Neural & 30.0 & 0.98 & 20.42 & \textbf{1.53} & \textbf{0.89} & \textbf{0.49} & \textbf{0.75} \\
\bottomrule
\end{tabular}
}
\label{tab:planning-results}
\end{table*}

We build our scene graph from camera images using the Khronos framework~\cite{schmid2024khronos}. At the lowest layer, the scene graph contains a semantically annotated mesh of the environment geometry, representing obstacles in the environment, which we convert into a parallelization-friendly CAPT point cloud~\cite{ramsey2024collision}, allowing us to perform collision checking efficiently using fine-grained parallelism.

The neural cost function in \cref{sec:method_scenegraph} is implemented as a multilayer perceptron (MLP)~\cite{rumelhart1986learning} with five fully connected layers of 256 units and ReLU activations. For the training dataset, we uniformly sample $50000$ robot-independent camera poses in Isaac Sim that keep the object in view. These viewpoints are not tied to a specific robot configuration and can be used with any high-\dof platform via forward kinematics. We then render the corresponding images and evaluate perception costs using the YOLOE~\cite{wang2025yoloe} model. The neural cost function is trained on these perception costs across diverse objects and humans from the COCO dataset~\cite{lin2014microsoft}, ensuring applicability to both real and simulated~experiments.

\subsection{Multi-Object Detection in a Simulated Office}
\label{sec:experiment_simulation}

We evaluate our approach in a simulated environment containing objects commonly found in an office, such as tables, chairs, humans, and monitors (an example is shown in ~\cref{fig:simulation_experiment}). This setting reflects typical scenarios faced by robotic assistants in office environments, where the robot must perform navigation or delivery tasks while monitoring multiple task-relevant objects, such as screens or humans. The task is to plan collision-free motions from a start to a goal while ensuring that the robot maintains visibility of monitors placed around the environment. To generate test cases, we sample $100$ motion planning problems by selecting random collision-free start and goal configurations on opposite sides of the room, ensuring that the robot must traverse the environment while balancing motion and perception costs. 
As illustrated in \cref{fig:simulation_experiment}, we place the objects of interest in physically plausible locations (e.g., resting on a surface rather than floating in the air) to create realistic and meaningful scenarios for evaluation.

We compare \mopsprm against three baselines: ``Closest-Object Low-\dof'', ``Closest-Object'', and ``Lowest-Cost-Object''. At a configuration $\bfq$,
``Closest-Object Low-\dof'' and ``Closest-Object'' always monitor the nearest object by projecting the camera view toward it, while ``Lowest-Cost-Object'' selects the object with the lowest perception cost as evaluated by the same neural cost function used in \mopsprm. 
In ``Closest-Object Low-\dof'', planning is restricted to the non-holonomic base with all other joints fixed, resembling perception-aware planning for aerial or ground robots. A comparison on the movement of camera joints is included in~\cref{fig:simulation_experiment}, with sample paths from the same environment. 
Both ``Closest-Object Low-\dof'' and ``Closest-Object'' define the perception cost as the distance to the selected object, whereas ``Lowest-Cost-Object'' instead uses the neural cost function of \mopsprm. We set time limits for PRM construction, adapted from OMPL~\cite{sucan2012open}, to ensure similar number of nodes across all methods: 5 seconds for ``Closest-Object Low-\dof'' and ``Closest-Object'', and 30 seconds for ``Lowest-Cost-Object'' and \mopsprm.

Perception performance is evaluated using YOLOE~\cite{wang2025yoloe} for object detection and Deep SORT~\cite{Wojke2017simple} for tracking, whose ``track rate" metric describes the benefits of continuously monitoring multiple objects beyond detection. One key metric is the average number of objects detected per frame, $\overline{D}$, which measures frame-to-frame visual coverage by averaging the number of detected objects across all frames along a trajectory. Detection confidence is measured in two forms: the average confidence $\overline{C}$, computed as the mean confidence score $s_i$ over all successful detections in the set $S$, and the scaled average confidence $\overline{C}_\mathrm{sc} = \overline{D}\,\overline{C}$, which emphasizes the ability to maintain both high-confidence detections and consistent multi-object coverage along the trajectory.

\cref{tab:planning-results} summarizes the results across the 100 planning problems. Both ``Closest-Object Low-\dof'' and ``Closest-Object'' incur lower computational overhead, as reflected in their significantly faster PRM construction and planning times. 
However, their strict focus on the nearest object leads to reduced coverage, evident from a lower average number of objects detected per frame $\overline{D}$. 
Their confidence metrics also lag behind, since they do not account for accurate perception cost estimates from each robot configuration. Meanwhile, the `Lowest-Cost-Object'' baseline achieves confidence scores comparable to \mopsprm by leveraging the neural cost function, but its inability to consider multiple objects simultaneously results in our method achieving more than $\sim 36\%$ improvement in the average number of detected objects per frame and a $\sim 17\%$ higher track rate. For this “track rate” metric, \mopsprm achieves the highest performance clearly surpassing all baselines. This highlights the advantage of continuously monitoring multiple objects beyond mere single-frame detection. While the baselines may achieve slightly shorter planning times or path lengths by focusing on an object at a time, \mopsprm explicitly accounts for multi-object monitoring, and hence, substantially improves perception performance.

\begin{figure}[t]
    \centering
    \includegraphics[width=\columnwidth]{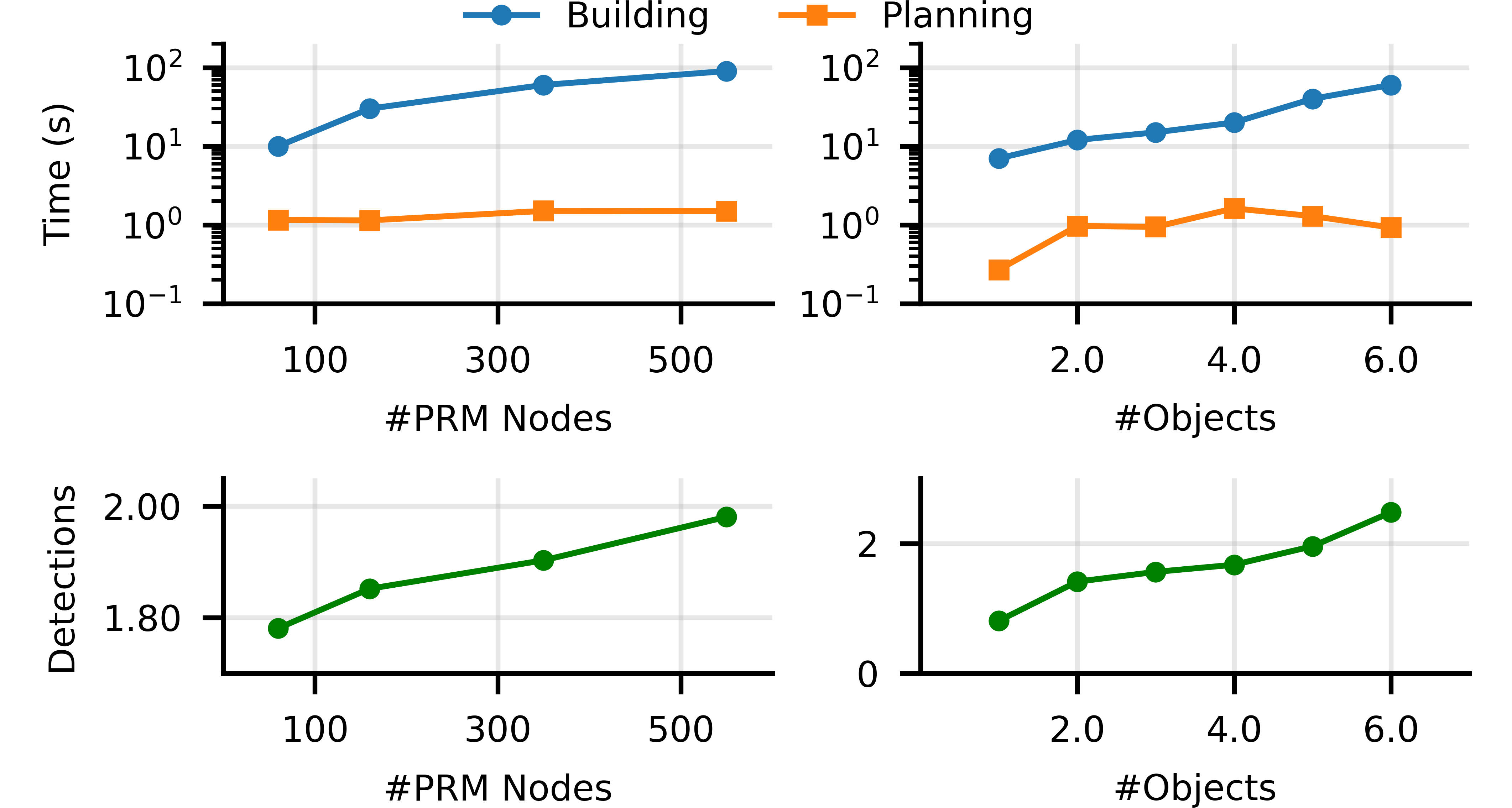}
    \caption{
    Performance of \mopsprm under varying number of objects and PRM sizes. In the first column, the number of objects is fixed at five. In the second, PRM size is approximately 300 nodes.  We report the planning and PRM construction times, and the average number of detections per frame.
    }
    \label{fig:prm_statistics_plot}
\end{figure}

\cref{fig:prm_statistics_plot} illustrates how our planner's performance scales with the PRM size and the number of objects. With the number of objects fixed at $5$, increasing the PRM size increases roadmap construction time, while planning time remains low, typically around or below one second. The average number of detections per frame also increases, indicating that a denser roadmap supports more stable perception quality. When the PRM size is fixed at roughly 300 nodes, adding more objects drives up construction time and modestly increases planning time. At the same time, perception performance improves as the number of PRM nodes or objects increases, as reflected in higher average detections per frame. While the PRM construction is time-consuming, it only occurs once, and can be reused multiple times for path generation. %leading to significantly lower planning time. 
The results suggest that our~approach remains practical as the problem size increases, with most of the overhead concentrated in the one-time construction stage.

\subsection{Real Robot Experiments}
\label{sec:experiment_real}
As shown in \cref{fig:real_robot_dynamic}, the robot is placed in an indoor environment and is tasked to move from a starting position shown in red to the corner of the room shown in green. To reach this goal, the robot must pass through a narrow passage created by an intervening chair, resulting in a challenging scenario with multi-modal solutions and high collision risks.  

Unlike the simulation experiments, this setup introduces a different challenge, since the objects are farther apart and facing different directions. While monitoring multiple objects simultaneously, the robot must transition its focus between different objects of interest while maintaining smooth motion and maximizing perception scores along the trajectory. 

The robot is tasked to consider the monitor near its starting position and a person near the end position, while detecting one of two objects, either a robot painting (painting 1) or a landscape painting (painting 2) placed on a cabinet, as shown in~\cref{fig:real_robot_dynamic}. The priority of monitoring each object is encoded by user-defined weights, as described in~\cref{sec:method_planner}.

\cref{fig:real_robot_dynamic} illustrates how \mopsprm generates trajectories based on which painting is prioritized. The two resulting paths are shown in yellow and blue, each highlighting a representative robot configuration along the path corresponding to the case where the respective painting is given higher weight.
Averaged over 100 runs, our planner takes around 1.64 seconds to generate each plan and takes around 30.0 seconds to build the PRM. The path length for this experiment, measured as the Euclidean distance at all 5-\dof
edges of the path, is around 19.13 for the yellow path and around 18.10 for the blue path. The average number of objects detected in each frame is around 0.84 on the yellow path and 0.81 on the blue path. This experiment demonstrates the planner’s ability to monitor multiple objects while respecting the assigned weights of each object of interest. 

\section{Discussion}
\label{sec:conclusion}
We develop \mopsprm, a roadmap-based perception-aware motion planner for high-\dof robots tasked with multi-object monitoring. Perception awareness is modeled via a costmap anchored to objects of interest in a scene graph, guiding the perception-aware PRM construction and A* search to produce paths that balance motion efficiency with perception quality. This enables applications such as museum patrol, patient monitoring, or industrial inspection where robots must move efficiently while maintaining visibility of key objects. As scene graphs have shown tremendous potential for task and motion planning in semantically rich environments, our work serves as a first step towards perception-aware task and motion planning for high-\dof robots, and can be further extended to leverage the scene graph's topology for perception-aware task-level reasoning. Besides, we also aim to extend the framework to include tree-based planners (e.g., RRT variants), handle map uncertainty for more robust planning, explore perception-aware motion planning in dynamic and interactive environments, and handle previously unseen objects, e.g., by incorporating open-set object detection~\cite{liu2024grounding}.
%\clearpage
%\printbibliography{}
\bibliographystyle{cls/IEEEtran}
\bibliography{bib/references}

\end{document}